\documentclass[conference]{IEEEtran}
\usepackage[utf8]{inputenc} 
\IEEEoverridecommandlockouts
\usepackage{cite}
\usepackage{amsmath,amssymb,amsfonts}
\usepackage{algorithmic}
\usepackage{graphicx}
\usepackage{textcomp}
\usepackage{xcolor}
\usepackage{url}
\usepackage[colorinlistoftodos]{todonotes}

\def\BibTeX{{\rm B\kern-.05em{\sc i\kern-.025em b}\kern-.08em
    T\kern-.1667em\lower.7ex\hbox{E}\kern-.125emX}}
\begin{document}

\title{Applying Genetic Programming to Improve Interpretability in Machine Learning Models
\thanks{This work has been supported by the Brazilian agencies (i) National Council for Scientific and Technological Development (CNPq); (ii) Coordination for the Improvement of Higher Education (CAPES) and (iii) Foundation for Research of the State of Minas Gerais (FAPEMIG, in Portuguese).}
\thanks{MINDS Laboratory -- \texttt{https://minds.eng.ufmg.br/}}
}

\author{\IEEEauthorblockN{ Leonardo Augusto Ferreira, Frederico Gadelha Guimarães}
\IEEEauthorblockA{
\textit{Machine Intelligence and Data Science (MINDS) Lab}\\
\textit{Department of Electrical Engineering  } \\
\textit{Universidade Federal de Minas Gerais, UFMG, }\\
31270-000 Belo
Horizonte – MG, Brazil \\
leauferreira@cpdee.ufmg.br, fredericoguimaraes@ufmg.br\\
ORCID 0000-0001-9238-8839}
\and
\IEEEauthorblockN{Rodrigo Silva}
\IEEEauthorblockA{\textit{Department of Computer Science} \\
\textit{Universidade Federal de Ouro Preto, UFOP}\\
35400-000 Ouro Preto - MG, Brazil \\
rodrigo.silva@ufop.edu.br \\
ORCID 0000-0003-2547-3835}
}

\maketitle

\begin{abstract}
Explainable Artificial Intelligence (or xAI) has become an important research topic in the fields of Machine Learning and Deep Learning. In this paper, we propose a Genetic Programming (GP) based approach, named Genetic Programming Explainer (GPX), to the problem of explaining decisions computed by AI systems. The method generates a noise set located in the neighborhood of the point of interest, whose prediction should be explained, and fits a local explanation model for the analyzed sample. The tree structure generated by GPX provides a comprehensible analytical, possibly non-linear, symbolic expression which reflects the local behavior of the complex model. We considered three machine learning techniques that can  be recognized as complex black-box models: Random Forest, Deep Neural Network and Support Vector Machine in twenty data sets for regression and classifications problems. Our results indicate that the GPX is able to produce more accurate understanding of complex models than the state of the art. The results validate the proposed approach as a novel way to deploy GP to improve interpretability.
\end{abstract}


\section{Introduction}
Advances in Machine Learning (ML)  and Deep Learning (DL) have had a profound impact in science and technology. These techniques have  had many recent successes, achieving unprecedented performance in tasks such as image classification, machine translation and speech recognition, to cite a few. The remarkable performance of Artificial Intelligence (AI) methods and the growing investment on AI worldwide will lead to an ever-increasing utilization of AI systems, having a significant impact on society and everyday life decisions.  However, depending on the model used, understanding why it makes a certain prediction can be difficult. This is particularly the case with the high performing DL models and ML models in general. The more complex the model, the more opaque its decisions are to human understanding. Black-box ML models are increasingly used in critical applications, leading to an urgent need for understanding and justifying their decisions. This difficulty is an important impediment for the adoption of ML, particularly DL, in domains such as healthcare \cite{pulmao2002}, criminal justice and finance \cite{survey}. The term Explainable AI (or XAI) has been adopted by the community to refer to techniques that help the understanding of decisions or results of AI artifacts by a given audience, which can be domain experts, regulatory agencies, managers, decision-makers, policy-makers or users affected by these decisions.


The interpretability problem can be understood as the technical challenge of explaining AI decisions, especially when the underlying AI technology is perceived as a black-box model. Humans tend to be less willing to accept decisions that are not directly interpretable and trustworthy \cite{Zhu2018}. Therefore, users should be able to understand the models outputs in order to trust them. 
There are two types of trusting, as described in \cite{lime}: one about the model and another about the prediction. Though similar, they are not the same thing. The first one is related to whether someone will choose a model or not, whereas the second relates to whether someone will make a decision relying on that prediction. 

Interpretability, as pointed out in \cite{xaiSurvey}, is associated with a human perception i.e. the ability to classify something or someone according to their main characteristics. This idea applied to ML models would be to highlight the main features that contributed to a prediction. Other works, such as \cite{hall2018} and \cite{molnar2019},  define interpretability as ``\textit{the ability to explain or to present in understandable terms to a human}''. In other words, a model can be defined as  explainable whether its decisions are easier for a human to understand. 

Another issue involving interpretability goes beyond trusting some model or prediction. The European Union has recently deployed the General Data Protection Regulation (GDPR) as pointed out in \cite{gdpr1} and \cite{gdpr2}. GDPR directly deals with subjects related to European citizens' data, for example: it prohibits judgments based solely on automated decisions \cite{inside_black_box}. European Union’s new GDPR has a major impact in deploying machine learning algorithms and AI-based systems. It restricts automated individual decision-making which ``significantly affects'' users. The law also creates a ``right to explanation,'' whereby someone has a right to be given an explanation for an output of the algorithm \cite{Goodman2017EuropeanUR}.

There are many important results of interpretability and explainability in the literature. For instance, interpretable mimic learning \cite{Che2016} is an approach in which the behavior of a slow and complex model is approximated by a faster and simpler model (also more transparent) with comparable performance. The idea is that by mimicking the performance of other complex models, one is able to compress the learning into a more interpretable model and derive rules or logical relationships. In this regard, it is possible to cite Lime (Local Interpretable Model-Agnostic Explanations)  \cite{lime} and SHAP (SHapley Additive exPlanations) \cite{shap}, which have been widely used for interpretability.
Despite their success, both Lime and SHAP assume that a linear model is a good local representation of the original one. This simplification may cause, in some circumstances, a significant loss in the mimicking model accuracy, which may spoil the final interpretation.

In this paper we present an approach to interpretability based on Genetic Programming (GP), named Genetic Programming Explainer (GPX). GP has the ability to produce linear and nonlinear models increasing the flexibility of the aforementioned methods. The evolution process of the GP algorithm ensures the selection of the best features for the analyzed sample. Moreover, the tree structure representation naturally provides an analytical expression that reflects a local explanation about the model's prediction. The main goal is to produce an accurate mimicking model which preserves the advantages of having a closed mathematical expression such as readability and the ability of computing partial derivatives to assess the sensitivity of the output with respect to the input parameters.
According to the taxonomy recently advocated by \cite{BARREDOARRIETA202082}, the proposed approach can be categorized as a model agnostic technique for post-hoc explainability, able to provide a local explanation for a specific prediction output given by a complex black-box model.

In this work we have considered the following pre-trained complex ML algorithms that can be recognized as black-box models: Random Forest \cite{random}, Supppot Vector Machines (SVM) and a Deep Neural Network (DNN)  \cite{rna} in a number of different data sets (10 regression and 10 binary classification problems available in public repositories). These methods have great performance in most ML problems, however, their explainability is low. For each pre-trained complex model, we compared  (GPX)  against other methods that could be used for generating local explanations: Lime and Decision Tree/Regression. The statistical analysis shows that GPX was able to better approximate the complex model, providing an interpretable explanation in terms of a symbolic expression. Genetic Programming  brought us a new approach for interpretability with a local explanation for black-box models predictions. In addition, we present two case studies to illustrate the proposed methodology and serve as a guide for future use. The first one is about predicting home prices in Boston area and the other one measures the progression of Diabetes over the years.

In summary, the proposed approach aims to contribute to improving interpretability of black-box ML by using automatic generation of model-agnostic explanations that are able to fit the input-output decisions performed by the more complex model, which would be otherwise impossible to be derived by usual analytical methods. The GP algorithm is applied locally and provides an analytical and visual explanation in terms of a human readable expression represented as a tree. These evolved explanations can be easily interpreted and help understanding these complex decisions.

This paper is organized as follows. Section \ref{sec:gp} reviews the main ideas of GP Algorithm and discusses how it will be applied for the purpose to provide interpretability. Section \ref{sec:interp} introduces some concepts about interpretability and describes our approach according to these. Section \ref{sec:Methodology} presents our methodology and the main idea of our solution for approaching the interpretability problem. Section \ref{sec:results} discusses the results of this article compared with the state of the art.

\section{Concepts of Interpretability } \label{sec:interp}

Humans are capable of making predictions about a subject and build a logical explanation to justify it. When a prediction is based on understandable choices, it gives the decision maker more confidence on the model \cite{survey}. On the other hand, Machine Learning and Deep Learning models are not able to provide the same level confidence. Their complexity and exorbitant number of parameters make them unintelligible for a human being and for most of the purposes they are seen as black-boxes. Humans tend to be resistant to techniques that are not well understood or that cannot be directly interpreted \cite{xaiSurvey}.

More recently, several strategies have been applied to understand how black-box models work. These strategies aim to decrease the opacity of artificial intelligence systems and to make these models more user friendly.
In order to address the opacity of some machine learning models, we first must introduce some concepts of interpretability \cite{molnar2019},\cite{survey}:

\begin{itemize}

\item \textbf{Comprehensibility:} refers to the ability of a learning algorithm to represent its learned knowledge in a human understandable fashion, in such a way that the knowledge representation can be comprehended.

\item \textbf{Interpretability:} It is defined as the ability to explain or to provide the meaning in understandable terms to a human. 

\item \textbf{Explainability:} is associated with the notion of explanation as an interface between humans and a decision maker that is, at the same time, both an accurate proxy of the decision maker and comprehensible to humans. 

\item \textbf{Transparency:} A model is considered to be transparent if by itself it is understandable. For instance, a decision tree is a transparent model. The model must allow the reproduction of every calculation step, it  needs to have a clear explanation of parameters and hyper-parameters and to provide an explanation about the learning process. Basically complex models can be considered as black-box models because they lack transparency.

\item \textbf{Functionality:} the model must provide an understandable output, it needs to have visualization tools and to ensure a local explanation \cite{lime}, \cite{shap}. The outputs must be presented in a user-friendly way.


\end{itemize}

\emph{Post hoc explainability} techniques employ other methods after training step to analyze a given model. Different techniques can be used to enhance  interpretability of a black-box model, such as text explanations, visual explanations, local explanations, explanations by example, explanations by simplification and feature relevance explanations \cite{xaiSurvey}.

More specifically, post hoc explainability by means of local explanations is an  approach applied for a prediction or set of predictions of pre-trained black-box models, for example: Lime \cite{lime} and SHAP \cite{shap} fall in this category. Lime generates a local explanation model while SHAP identifies feature relevance explanations.

The approach presented in this paper can be considered as a post hoc technique since the GP algorithm is applied to a pre-trained model. The goal of the GP is to produce a local symbolic model to provide a visual explanation in terms of a human readable expression represented as a tree. Thus, in this sense, the proposed method generates a knowledge representation that can be comprehended, a local explanation model that is interpretable, transparent and functional. 


Explainability and interpretability are extremely relevant issues nowadays. They are fundamental pieces for dealing with several philosophical and ethical aspects of the interaction between humans and AI artifacts.

\section{Evolving Explanations} \label{sec:Methodology}

\subsection{Genetic Programming} \label{sec:gp}

\begin{figure}[!htb]
\centerline{\includegraphics[width=.42\textwidth]{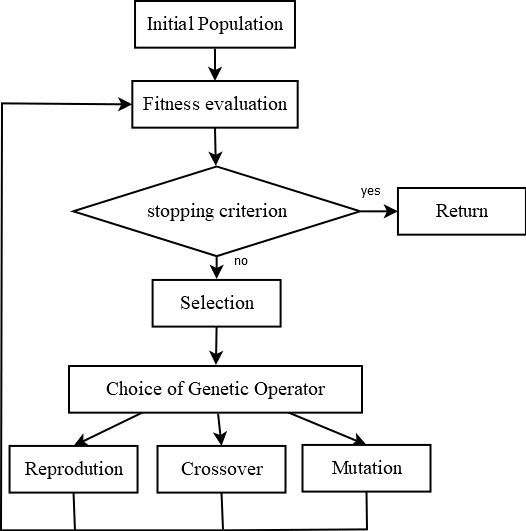}}
\caption{This flowchart presents the steps of the evolutionary process in GP.}
\label{fig:gp}
\end{figure}

Genetic Programming (GP) was developed by Koza \cite{koza1992} to evolve functional structures such as algebraic expressions, computer programs, and logical expressions \cite{manual}. 
GP works with a population of programs, usually initialized at random. In a GP there is a problem to be solved and the fitness of the individuals is related to how well they solve this problem. Thus, the GP algorithm evolves this population until it finds the best solution for the problem. Figure \ref{fig:gp} presents all steps of the GP algorithm.

The GP algorithm is relatively simple to describe, as observed in Figure \ref{fig:gp}. The first step is to generate a random population with several individuals. The next step is to evaluate the fitness of each individual to measure how well an individual solves the problem at hand. If some individual satisfies the stopping criterion, the algorithm ends. Otherwise, it goes to the selection step in which the algorithm will favor the better individuals based on fitness evaluation. After the selection step a genetic operator is chosen randomly to generate new individuals \cite{koza1992, manual} . 

\begin{figure}[!h]
\center
\includegraphics[width=.18\textwidth]{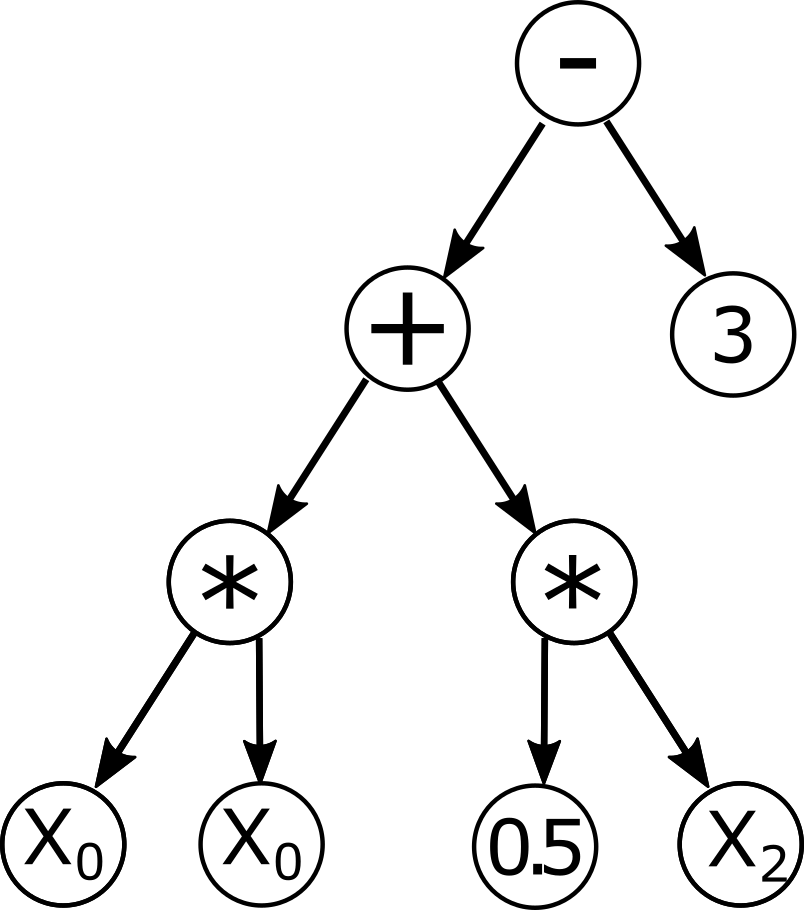}
\caption{This binary tree represents the expression: $f(x)=x_0^2 + 0.5x_2 - 3$. The terminal nodes are constants or input variables and the internal nodes are the functions or operations. }
\label{fig:tree}
\end{figure}

In this work, genetic programming is used to evolve nonlinear symbolic expressions for regression and classification problems. These expressions are represented in the program as a tree over which the genetic operations of mutation, crossover and reproduction are defined. Figure~\ref{fig:tree} illustrates the tree representation of a mathematical expression. 

The \textit{gplearn}\footnote{\texttt{https://gplearn.readthedocs.io/}} Python library was used as the GP search engine. This library extends from scikit-learn, a widely known Python library for ML.


\subsection{GP approach to local explanation} \label{sec:gp_approach}

In this section, the proposed approach to the interpretability problem is described. The example discussed here helps understanding the steps needed to achieve our local interpretation. 

Let $\mathbf{x}\in\mathbb{R}^n$ be an input fed to a complex pre-trained machine learning model. The first step in our method is to generate $m$ sample points around the input $\mathbf{x}$. This set of samples, called noise set, $\eta$, is created by sampling from a multivariate Gaussian distribution centered at $\mathbf{x}$ with covariance matrix, $\mathbf{\Sigma} = I_n \times \sigma$ where, $I_n$ is the $n \times n$ identity matrix and $\sigma$ is measured on training data.

The goal of GPX is to find the function, $f^* : \mathbb{R}^{n} \rightarrow \mathbb{R}$ which is easy to interpret and, at the same time, mimics the behavior of original complex model, $g : \mathbb{R}^{n} \rightarrow \mathbb{R}$, over the sample set, $\eta$. Formally, the problem to be solved by GPX can be defined as follows:   

\begin{equation}
f^* = \underset{f \in F, \mathbf{s}_i \in \eta}{\arg\min} \; d([f(\mathbf{s}_1),\hdots,f(\mathbf{s}_m)] -  [g(\mathbf{s}_1),\hdots,g(\mathbf{s}_m)])
\label{eq:argmin}
\end{equation}
where:

\begin{itemize}
\item $F$: set of all admissible functions $f : \mathbb{R}^{d} \rightarrow \mathbb{R}$;
\item $\{\mathbf{s}_1, \mathbf{s}_2, ..., \mathbf{s}_m\}$: samples form the noise set, $\eta$;
\item $g(\mathbf{s}_i)$: is the prediction given by the complex ML model;
\item $f(\mathbf{s}_i)$ is the prediction generated by a given individual in the population; and
\item $d(\cdot)$: some distance metric, usually the $l_2$-norm or the root-mean-square error (RMSE).
\end{itemize}


\begin{figure}[htb]
\centerline{\includegraphics[width=.35\textwidth]{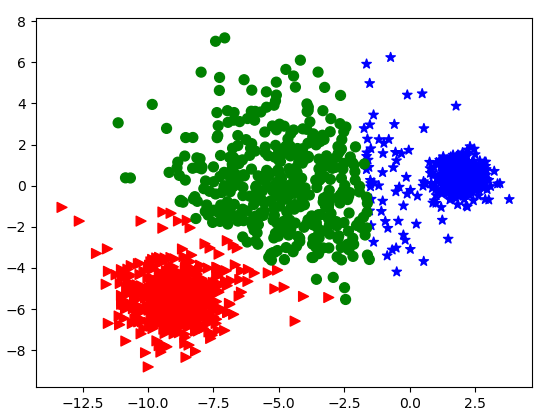}}
\caption{Suppose a classification problem in which there are 1,500 samples in the training data set in 3 separate classes. Class 1 are triangles, class 2 circles and class 3 stars. This data set was created by scikit-learn tool, where the standard deviation in each class is 1.0, 2.5, 0.5 respectively.}
\label{fig:3classe}
\end{figure}

As an example, Figure \ref{fig:3classe} illustrates some training data $X_{k \times n}$, where $k = 1,500$ samples and $n = 2$. Assume that a model, $g(\cdot)$, has been previously obtained from this data set.

\begin{figure}[htb]
\centerline{\includegraphics[width=.35\textwidth]{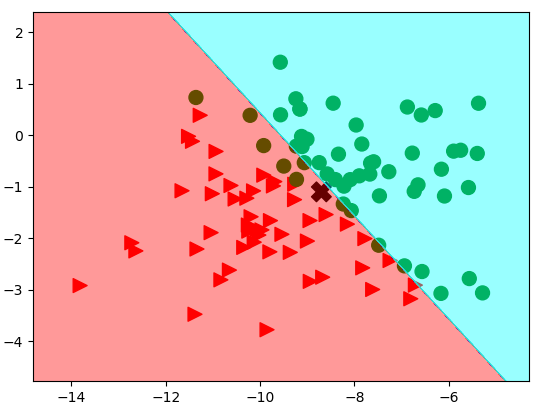}}
\caption{The bold X represents the input $\mathbf{x}$ for a complex ML model, whose prediction should be interpreted. The set $\eta$ contains 100 samples classified by the complex model. This figure also presents the surface decision generated by GP algorithm.}
\label{fig:my_distribution}
\end{figure}

After defining $\sigma$, it is possible to generate the noise set, $\eta$, located in the neighborhood of the point of interest $\mathbf{x}$ whose prediction should be explained. Figure \ref{fig:my_distribution} shows $\mathbf{x}$ and the noise set, $\eta$, generated randomly around $\mathbf{x}$. 

The next step is to apply GP to try to find $f^*$ as defined in Equation \eqref{eq:argmin}.

\begin{figure}[!htb]
\centerline{\includegraphics[width=.24\textwidth]{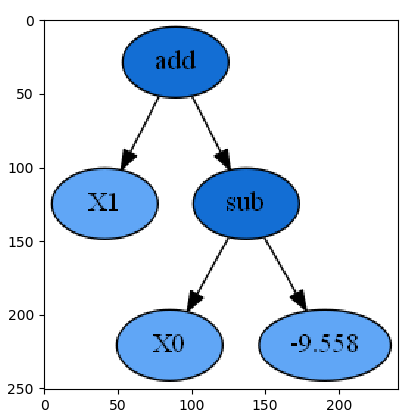}}
\caption{A tree structure representing an algebraic expression, $f(s) = x_1 + (x_0 - (-9.558))$. This structure was generated by GPX and visualized via \textit{graphviz}.}
\label{fig:tree_expression}
\end{figure}

Figure \ref{fig:tree_expression} illustrates the best representation found for $f^*$ by the GP. For didactic reasons, this example was built with two dimensions only and the produced representation for $f^*$ contains all the input features. It is important to highlight, however, that the GP is not obliged to use all the features. Thus, the produced model already indicates which variables are important. Besides, the returned closed mathematical expression allows the user to compute partial derivatives to assess the local importance of each feature.  

\section{Experimental Methodology} \label{sec:experiments}

\subsection{Data Sets} \label{sec:datasets}

In this work twenty data sets were used. Ten for classification and ten regression. All data sets were extracted from well-known repositories such as \textit{UCI Machine Learning Repository}\footnote{\texttt{https://archive.ics.uci.edu//}},  \textit{OpenML}\footnote{\texttt{https://www.openml.org}} and \textit{Kaggle}\footnote{\texttt{https://www.kaggle.com/}}. Our  selection was based on features variability between data sets and number of downloads. Table~\ref{tab:datasets_classficiations} lists the chosen data sets for the classification problems and the corresponding number of used features.

\begin{table}[htbp]
\caption{Classification Data Sets}
\begin{center}
\begin{tabular}{|l|c|}
\hline 

\textbf{Data Set} &  \textbf{Features} \\ 
\hline 
Diabetes \cite{diabetes}& 8 \\ 
\hline 
Steel Plates Faults \cite{steel_2010} &  33  \\ 
\hline 
The Monk's Problems 2 \cite{monks2}  &  6  \\ 
\hline 
Phoneme \cite{phoneme} &  5  \\ 
\hline 
Blood Transfusion Service Center \cite{blood} &  4   \\ 
\hline 
Ozone Level Detection \cite{ozone} &  72   \\ 
\hline 
Hill-Valley \cite{uci:2019} &  100   \\ 
\hline 
EEG Eye State \cite{uci:2019} &  14   \\ 
\hline 
Spambase \cite{uci:2019} &  57   \\ 
\hline 
ILPD (Indian Liver Patient Data Set) \cite{ilpd} &   10  \\ 
\hline 

\end{tabular} 
\label{tab:datasets_classficiations}
\end{center}
\end{table}

Table \ref{tab:datasets_regression} presents the data sets chosen for the regression problems and the corresponding number of used features.

\begin{table}[htbp]
\caption{Regression Data Sets}
\begin{center}
\begin{tabular}{|l|c|}
\hline 

\textbf{Data Set} &  \textbf{Features} \\ 
\hline
Diabetes \cite{uci:2019} &  10  \\ 
\hline 
Boston \cite{sklearn_api} &  13  \\ 
\hline 
Fetch California Housing \cite{sklearn_api} &  8  \\ 
\hline 
Bike Sharing in Washington D.C.  &  13  \\ 
\hline 
Red Wine Quality \cite{red_wine} &  11  \\ 
\hline 
House Sales in King County, USA &  18  \\ 
\hline 
GPU Kernel Performance \cite{gpu_kernel} &  14  \\ 
\hline 
Beer Consumption - São Paulo &  5  \\ 
\hline 
Houses to rent data &  8 \\ 
\hline 
Predicting Compressive Strength of Concrete &  10  \\ 
\hline 
\end{tabular} 
\label{tab:datasets_regression}
\end{center}
\end{table}

\subsection{Complex, black-box, models} \label{sec:black_box}

There are several machine learning models which can be considered black-box. In this work we have selected three of the most popular to serve as ``complex models''. That is, model for which explanations have to be produced. These models are the Random Forest, Deep Neural Networks and Suport Vector Machines. They are briefly introduced below. 

Random Forest is a type of tree ensemble which have shown improvements for the generalization of a single decision tree and has achieved remarkable results in the literature \cite{random,breiman2001random}. Our experiments set Random Forest with up two thousand trees. The more trees are used in the architecture the more complexity is brought to the understanding of the model.

Deep neural networks (DNNs) \cite{rna} are artificial neural networks (ANNs) with multiple layers between the input and output layers. The multi-layer architecture is inspired in the human brain structure and has shown to be quite effective in many difficult classification and regression problems \cite{Goodfellow2015DeepL}. This architecture, however, represents compositions of non-linear functions which are not easy to interpret.   

Finally, Support Vector Machines (SVMs) \cite{vapnik} are classifiers which find a separating hyperplane, such that the distance on either side of that hyperplane to the next-closest data points is maximized. In other words, given labeled training data, the algorithm outputs an optimal hyperplane which categorizes new examples. Support Vector Regresssion (SVR) is the extension of SVMs for regression problems \cite{drucker1997support}. The SVR tries to fit the error within a certain threshold and can also be considered opaque since it is hard to interpret its decisions, specially when nonlinear kernel functions are used.

\subsection{Competing Explainers}

In this section we present the competing explainers. Just like our method, the idea is to sample a noise set, $\eta$, around the point of interest, $\mathbf{x}$. The set $\eta$ is then used to build a local comprehensible model which will be used as an explainer. We say that an explainer, $f$, ``understands'' the complex model, $g$, if its error on the noise set, $\eta$, relative to the complex model predictions is small. For regression problems, this concept may be translated into the Root Mean Squared Error between the model and the explainer as follows: 

\begin{equation}\label{eq:understand_regression}
    u_r(f) =   \frac{1}{|\eta|} \sum_{\mathbf{s}_i \in \eta} (f(\mathbf{s}_i) - g(\mathbf{s}_i) )^2
\end{equation}

For classification problems, it may be formalized as the accuracy of the explainer with respect to the complex models predictions.

\begin{equation}\label{eq:understand_classification}
     u_c(f) =   \frac{1}{|\eta|} \sum_{\mathbf{s}_i \in \eta} h(\mathbf{s_i}) 
\end{equation}

\noindent where,

\begin{equation}
    h(\mathbf{s}_i) = \begin{cases}
    1 & \text{ if } f(\mathbf{s}_i) = g(\mathbf{s}_i)\\ 
    0 & \text{ if } f(\mathbf{s}_i) \neq g(\mathbf{s}_i) 
    \end{cases} 
\end{equation}




The first method, named Lime \cite{lime}, generates an explainer based on a linear least squares method with $l_2$-norm regularization \cite{Tikhonov:1963}. This linear model is used into Lime in order to measure locally the feature importance. 

Another good candidate for explainer is the Decision Tree (DT). A DT is, as the name implies, a tree where each node represents a feature, each branch represents a decision, and each leaf represents a prediction. By making a path from the root to the prediction leaf node, the user can obtain an explanation for that prediction.  

The performance of Lime, DT and GPX (explained in section \ref{sec:gp}) as explainers of Support Vector Machines, Neural Networks and Random Forests are discussed next. 

\section{Testing Explainers Accuracy} \label{sec:results}

\subsection{Experimental Setup}

In this section we test the ability of Lime, GPX, DTs to understand the different complex models discussed in section \ref{sec:black_box}. The experimental steps can be described as follows:

\begin{enumerate}
    \item Divide all the data sets described in section~\ref{sec:datasets} into training and test data. We used 80\% for training and 20\% for test.
    \item Train the complex models (described in section \ref{sec:black_box}) with training data. 
    \item Using the trained complex models, predict the  value of 100 random samples selected from the test set. At this point, we have six thousand predictions for the regression and classification problems together (20 data sets $\times$ 100 predictions $\times$ 3 complex models).
    \item Build the three explainers (Lime, GPX, DT) for each of these six thousand predictions in order to measure which interpreter can better understand the black-box prediction. At this time we have 18,000 data in order to apply a statistical analysis.
\end{enumerate}


The GPX hyper-parameters were set as follows:
\begin{itemize}
    \item Population size: 100 individuals.
    \item Probability of crossover: 70\%.
    \item Probability of hoist mutation: 5\%. This mutation is called hoist mutation because the method chooses randomly a subtree and hoist it into the tree. 
    \item Probability of point mutation: 10\%. This mutation selects a random node to be replaced. 
    \item Search interval: $[-100, 100]$.
\end{itemize}

The other hyper-parameters are set to the library default values.

The noise set, $\eta$ (see section \ref{sec:gp_approach}), has the same one thousand samples for each explainer. Thus, the explainers always access same data. In order to measure how well the explainers understand the complex models  \eqref{eq:understand_regression} and  \eqref{eq:understand_classification} are used for the regression and classification problems, respectively.

\subsection{Results}


Table \ref{tab:exp_error} presents the average and the standard deviation of the error, computed with \eqref{eq:understand_regression}, for each explainer across the regression problems.

\begin{table}[!h]
    \centering
    \caption{Explainers error}
    \begin{tabular}{c|c|c}
        Explainer & Average Error & Error Standard Deviation  \\
         \hline
         DT     & 0.083	  & 0.329 \\
         GPX    & 0.065   & 0.508 \\
         Lime   & 7.577   & 36.913
    \end{tabular}
    \label{tab:exp_error}
\end{table}

A Permutation Pairwise Test\footnote{See \url{https://rdrr.io/cran/rcompanion/man/pairwisePermutationTest.html}} was performed in order to test the hypothesis that the difference between the error means is zero. The results are shown in Table \ref{tab:paiwise_error} where P.adjust represents the P-Value adjusted with the Bonferroni Method. Considering a confidence level of $95\%$, it is possible to say that there is no difference in the mean error presented by the DT and the GPX. On the other hand, the results indicate that both DT and GPX, understand the complex model better than Lime.      

\begin{table}[!ht]
    \centering
        \caption{Pairwise Permutation Test for the Error Difference}
    \begin{tabular}{c|c|c|c}
        Comparison     & Stat   & P-value   & P.adjust  \\
        \hline
        DT - GPX  = 0  & 1.599  &  0.1099   & 0.3297 \\
        DT - Lime = 0  & -11.01 & 3.515e-28 & 1.054e-27 \\
        GPX - Lime = 0 & -11.03 & 2.668e-28 & 8.004e-28 \\
    \end{tabular}

    \label{tab:paiwise_error}
\end{table}


Table \ref{tab:exp_acc} presents the average and the standard deviation of the accuracy, computed with \eqref{eq:understand_classification}, for each explainer across the classification problems.


\begin{table}[!h]
    \centering
        \caption{Explainers Accuracy}
    \begin{tabular}{c|c|c}
        Explainer & Average Accuracy & Accuracy Standard Deviation  \\
         \hline
         DT     & 0.852	& 0.104 \\
         GPX    & 0.899	& 0.130 \\
         Lime   & 0.658	& 0.334
    \end{tabular}

    \label{tab:exp_acc}
\end{table}

A Permutation Pairwise Test was performed again in order to test the hypothesis that the difference between the accuracy means is zero. The results are shown in Table \ref{tab:paiwise_acc} where P.adjust represents the P-Value adjusted with the Bonferroni Method. Considering a confidence level of $95\%$, the results indicate that, in average, GPX was better than both DT and Lime in understanding the complex models. The DT, in turn, was better than Lime.       
\begin{table}[!ht]
    \centering
        \caption{Pairwise Permutation Test for the Accuracy Difference}
    \begin{tabular}{c|c|c|c}
    Comparison     &Stat          & P-value   & P.adjust \\
    \hline
    DT - GPX = 0   &-15.09        & 1.866e-51 & 5.598e-51\\
    DT - Lime = 0  &28.31         & 0         & 0.000e+00\\
    GPX -Lime = 0  &33.27         & 0         & 0.000e+00\\
    \end{tabular}

    \label{tab:paiwise_acc}
\end{table}

The complete table of results as well as the analysis scripts are available at \url{https://github.com/leauferreira/GpX}.

Overall, these results show that, at least for the scenarios presented here, the Lime assumption that the complex model can be locally approximated by a linear model, is not always adequate. The DT and the GPX specially were superior methods in understanding the complex model.    

Having shown that the proposed methodology can indeed produce accurate local models, in the next section, a demonstration of the use of GPX in practice is presented. 



\section{Case study on interpretability for Random Forest Regressor 
} \label{sec:case_study}

In this case study we selected the Boston and the Diabetes data sets, see Table \ref{tab:datasets_regression}, and the \textit{Random Forest Regressor} with 2,000 estimators (trees),  from \textit{scikit-learn}. Both data sets (Diabetes and Boston) can be found in \textit{scikit-learn}

The Boston data set has 13 features, 506 samples and the target consists of home prices in Boston. The data set was randomly split into training and test sets with $80\%$ of the samples for training and the remainder for testing. Apart from the number of trees the other hyper-parameters were set to the default values in the library. It is hard to analyze the joint decision performed by 2,000 trees, therefore explainability of Random Forest is low, although its performance is high.

After the training step we take the first sample in the test set, $x_1$, and apply the process described in Section \ref{sec:gp_approach} in order to create a noise around $x_1$. Then, the GP algorithm is trained with the noise set, $\eta = \{ s_1,...,s_n \}$ where the targets are given by $g(s_i)$, $\forall  s_i \in \eta$, and $g$ is the  \textit{Random Forest Regressor} model. 

\begin{figure}[htb]
\centerline{\includegraphics[width=.5\textwidth]{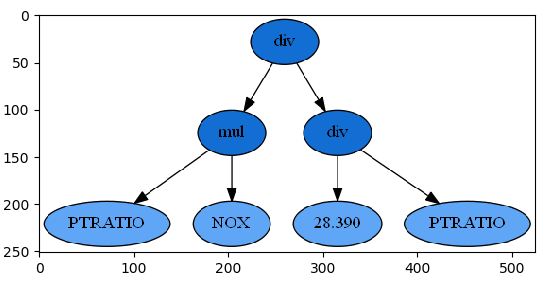}}
\caption{GP algorithm output for the \textit{Random Forest Regressor} prediction applied to instance $x_1$ in Boston data set.}
\label{fig:instance}
\end{figure}

The Boston data set consists of $d = 13$ features. However, as observed in Figure \ref{fig:instance}, after the evolution process only two features were chosen by the GP. They were:

\begin{itemize}
    \item \textbf{PTRATIO }: pupil-teacher ratio by town
    \item \textbf{NOX}: nitric oxides concentration (parts per 10 million)
\end{itemize}

In  \eqref{eq:boston_instance1} we changed \textbf{PTRATIO} to $x_{ptratio}$ and \textbf{NOX} to $x_{nox}$. The tree structure in Figure \ref{fig:instance} represents the following equation:

\begin{equation}\label{eq:boston_instance1}
    f^*(s) = \frac{x_{ptratio}^2 x_{nox}}{28.390}
\end{equation}

\begin{figure}[htb]
\centerline{\includegraphics[width=.34\textwidth]{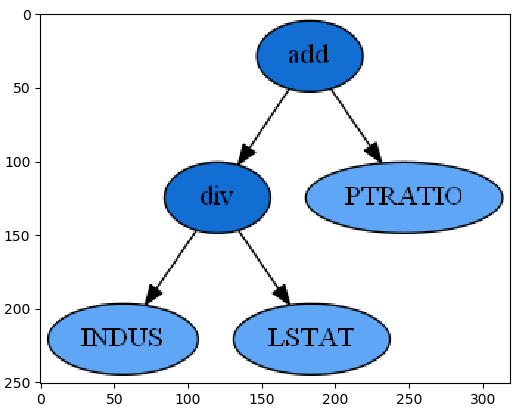}}
\caption{GP algorithm output for the \textit{Random Forest Regressor} prediction applied to instance $x_2$ in Boston data set.}
\label{fig:instance2}
\end{figure}

Figure \ref{fig:instance2} presents the result of the same process but in  a different area, by considering sample $x_2$. It is possible to observe that different  features were chosen during the evolutionary process.
The features chosen for the solution shown in Figure \ref{fig:instance2} were:
\begin{itemize}
    \item \textbf{PTRATIO:} pupil-teacher ratio by town
    \item \textbf{INDUS} proportion of non-retail business acres per town centres
    \item \textbf{LSTAT}  lower status of the population
\end{itemize}

The expression represented in Figure \ref{fig:instance2} can be defined as:

\begin{equation} \label{eq:bostont_instance2}
f^*(s) = \frac{x_{indus}}{x_{lstat}} + x_{ptratio}
\end{equation}

Based on the equations \eqref{eq:boston_instance1} and \eqref{eq:bostont_instance2} we can understand the behavior around $x_1$ and $x_2$. First of all, it is possible to observe that the feature PTRATIO (pupil-teacher ratio by town) is relevant to define home prices in the neighborhood of both instances ($x_1, x_2$). However, in the neighborhood of $x_2$, PTRATIO has more influence.  

These results can help a decision-maker to understand which features contribute the most to the increase in the price in a neighborhood. Moreover, it is possible to know which features changed in order to increase or decrease home prices. In this regard, it is useful to compute the gradient of the model output with respect to the selected input parameters. Lets take equation \eqref{eq:boston_instance1} as an example. Its gradient is given by:

\begin{equation} \label{eq:grad_instance1}
    \nabla f^*(s) = \left [ \frac{x_{ptratio}x_{nox}}{14.195},  \frac{x_{ptratio}^2}{28.390} \right ]^T
\end{equation}

By analyzing equations \eqref{eq:boston_instance1} and \eqref{eq:grad_instance1} it is easy to understand how and by how much each feature affects housing prices. 

\begin{figure}[htb]
\centerline{\includegraphics[width=.4\textwidth]{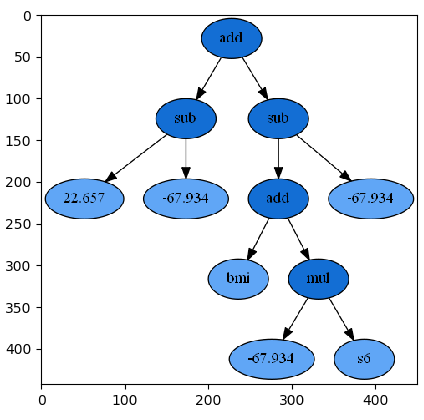}}
\caption{GP algorithm output for the \textit{Random Forest Regressor} prediction applied to instance $x_{d1}$ in Diabetes data set.}
\label{fig:instance1_diabetes}
\end{figure}

The Random Forest Regressor has also been applied to the Diabetes data set. This data set consists of 402 samples and 10 features. In this case, the target is a quantitative measure of disease progression. The GPX was set with the same hyper-parameters used for the Boston data set.  Figure \ref{fig:instance1_diabetes} presents the result for the Diabetes data set.

The evolutionary process chose a final tree with two features:

\begin{itemize}
    \item \textbf{bmi}: Body mass index
    \item \textbf{$\mathbf{S_6}$:} blood serum measurements
\end{itemize}

The tree structure in Figure \ref{fig:instance1_diabetes} represents the following expression:

\begin{equation}\label{eq:random_diabetes}
    -67.934 x_{S_6} + x_{bmi} + 158.525 
\end{equation}
where $x_{bmi}$ is \textbf{bmi} and  $x_{S_6}$ is $\mathbf{S_6}$.

This expression shows for instance that decreasing body mass index is a relevant feature to decrease diabetes progression for that specific patient.

\section{Conclusion}

This paper presented an approach to the interpretability problem based on Genetic Programming. We discussed several concepts of interpretability and why this subject is relevant nowadays. 

The GP algorithm used was able to produce a non-linear algebraic expression as output, which in turn provides many opportunities for the interpretability of the more sophisticated ML algorithms. It naturally selects the most important features to build a local explanation around a given sample. Besides, the produced analytic expression allows for easy differentiation, which gives the sensitivity of the output with respect to each feature.  

Examples using the classic Boston and Diabetes data sets show that the proposed approach can be seen as a source for interpretability and help the decision maker to understand better how the complex model is making a decision. 

We submit the explainers GPX, Lime and Decision Tree to a stress test in order to measure which one can, locally, better understand the black-box model. The statistical  analysis  showed us that GPX, besides bringing a new approach to interpretability, presented better or at least similar results when compared with the state of the art. 



\bibliographystyle{IEEEtran}
\bibliography{ref}

\begin{thebibliography}{10}
\providecommand{\url}[1]{#1}
\csname url@samestyle\endcsname
\providecommand{\newblock}{\relax}
\providecommand{\bibinfo}[2]{#2}
\providecommand{\BIBentrySTDinterwordspacing}{\spaceskip=0pt\relax}
\providecommand{\BIBentryALTinterwordstretchfactor}{4}
\providecommand{\BIBentryALTinterwordspacing}{\spaceskip=\fontdimen2\font plus
\BIBentryALTinterwordstretchfactor\fontdimen3\font minus
  \fontdimen4\font\relax}
\providecommand{\BIBforeignlanguage}[2]{{%
\expandafter\ifx\csname l@#1\endcsname\relax
\typeout{** WARNING: IEEEtran.bst: No hyphenation pattern has been}%
\typeout{** loaded for the language `#1'. Using the pattern for}%
\typeout{** the default language instead.}%
\else
\language=\csname l@#1\endcsname
\fi
#2}}
\providecommand{\BIBdecl}{\relax}
\BIBdecl

\bibitem{pulmao2002}
\BIBentryALTinterwordspacing
Z.-H. Zhou, Y.~Jiang, Y.-B. Yang, and S.-F. Chen, ``Lung cancer cell
  identification based on artificial neural network ensembles,''
  \emph{Artificial Intelligence in Medicine}, vol.~24, no.~1, pp. 25 -- 36,
  2002. [Online]. Available:
  \url{http://www.sciencedirect.com/science/article/pii/S093336570100094X}
\BIBentrySTDinterwordspacing

\bibitem{survey}
S.~{Chakraborty}, R.~{Tomsett}, R.~{Raghavendra}, D.~{Harborne}, M.~{Alzantot},
  F.~{Cerutti}, M.~{Srivastava}, A.~{Preece}, S.~{Julier}, R.~M. {Rao}, T.~D.
  {Kelley}, D.~{Braines}, M.~{Sensoy}, C.~J. {Willis}, and P.~{Gurram},
  ``Interpretability of deep learning models: A survey of results,'' pp. 1--6,
  Aug 2017.

\bibitem{Zhu2018}
J.~{Zhu}, A.~{Liapis}, S.~{Risi}, R.~{Bidarra}, and G.~M. {Youngblood},
  ``Explainable ai for designers: A human-centered perspective on
  mixed-initiative co-creation,'' in \emph{2018 IEEE Conference on
  Computational Intelligence and Games (CIG)}, 2018, pp. 1--8.

\bibitem{lime}
\BIBentryALTinterwordspacing
M.~T. Ribeiro, S.~Singh, and C.~Guestrin, ``“{W}hy {S}hould {I} {T}rust
  {Y}ou?”: {E}xplaining the {P}redictions of {A}ny {C}lassifier,'' in
  \emph{Proceedings of the 22nd ACM SIGKDD International Conference on
  Knowledge Discovery and Data Mining}, ser. KDD ’16.\hskip 1em plus 0.5em
  minus 0.4em\relax New York, NY, USA: Association for Computing Machinery,
  2016, p. 1135–1144. [Online]. Available:
  \url{https://doi.org/10.1145/2939672.2939778}
\BIBentrySTDinterwordspacing

\bibitem{xaiSurvey}
E.~Tjoa and C.~Guan, ``A survey on explainable artificial intelligence (xai):
  Towards medical xai,'' 10 2019.

\bibitem{hall2018}
P.~Hall and N.~Gill, \emph{An Introduction to Machine Learning
  Interpretability: An Applied Perspective on Fairness, Accountability,
  Transparency, and Explainable AI}.\hskip 1em plus 0.5em minus 0.4em\relax
  O'Reilly Media, 2018.

\bibitem{molnar2019}
C.~Molnar, \emph{Interpretable Machine Learning. A Guide for Making Black Box
  Models Explainable}, 2019,
  \url{https://christophm.github.io/interpretable-ml-book/}.

\bibitem{gdpr1}
\BIBentryALTinterwordspacing
C.~Tankard, ``What the gdpr means for businesses,'' \emph{Network Security},
  vol. 2016, no.~6, pp. 5 -- 8, 2016. [Online]. Available:
  \url{http://www.sciencedirect.com/science/article/pii/S1353485816300563}
\BIBentrySTDinterwordspacing

\bibitem{gdpr2}
\BIBentryALTinterwordspacing
J.~P. Albrecht, ``How the gdpr will change the world,'' \emph{European Data
  Protection Law Review}, vol.~2, no.~3, 2016. [Online]. Available:
  \url{https://doi.org/10.21552/EDPL/2016/3/4}
\BIBentrySTDinterwordspacing

\bibitem{inside_black_box}
\BIBentryALTinterwordspacing
B.~P. Evans, B.~Xue, and M.~Zhang, ``What’s inside the black-box? a genetic
  programming method for interpreting complex machine learning models,'' p.
  1012–1020, 2019. [Online]. Available:
  \url{https://doi.org/10.1145/3321707.3321726}
\BIBentrySTDinterwordspacing

\bibitem{Goodman2017EuropeanUR}
B.~Goodman and S.~Flaxman, ``European union regulations on algorithmic
  decision-making and a "right to explanation",'' \emph{AI Magazine}, vol.~38,
  pp. 50--57, 2017.

\bibitem{Che2016}
Z.~Che, S.~Purushotham, R.~G. Khemani, and Y.~Liu, ``Interpretable deep models
  for icu outcome prediction,'' \emph{AMIA ... Annual Symposium proceedings.
  AMIA Symposium}, vol. 2016, pp. 371--380, 2016.

\bibitem{shap}
\BIBentryALTinterwordspacing
S.~M. Lundberg and S.-I. Lee, ``A unified approach to interpreting model
  predictions,'' pp. 4765--4774, 2017. [Online]. Available:
  \url{http://papers.nips.cc/paper/7062-a-unified-approach-to-interpreting-model-predictions.pdf}
\BIBentrySTDinterwordspacing

\bibitem{BARREDOARRIETA202082}
\BIBentryALTinterwordspacing
A.~B. Arrieta], N.~Díaz-Rodríguez, J.~D. Ser], A.~Bennetot, S.~Tabik,
  A.~Barbado, S.~Garcia, S.~Gil-Lopez, D.~Molina, R.~Benjamins, R.~Chatila, and
  F.~Herrera, ``Explainable artificial intelligence (xai): Concepts,
  taxonomies, opportunities and challenges toward responsible ai,''
  \emph{Information Fusion}, vol.~58, pp. 82 -- 115, 2020. [Online]. Available:
  \url{http://www.sciencedirect.com/science/article/pii/S1566253519308103}
\BIBentrySTDinterwordspacing

\bibitem{random}
F.~Livingston, ``Implementation of breiman’s random forest machine learning
  algorithm,'' \emph{ECE591Q Machine Learning Journal Paper}, pp. 1--13, 2005.

\bibitem{rna}
S.~Haykin, \emph{Neural Networks: A Comprehensive Foundation}, 2nd~ed.\hskip
  1em plus 0.5em minus 0.4em\relax USA: Prentice Hall PTR, 1998.

\bibitem{koza1992}
J.~R. Koza, \emph{Genetic Programming: On the Programming of Computers by Means
  of Natural Selection}.\hskip 1em plus 0.5em minus 0.4em\relax Cambridge, MA,
  USA: MIT Press, 1992.

\bibitem{manual}
\BIBentryALTinterwordspacing
A.~Gaspar-Cunha, R.~Takahashi, and C.~Antunes, \emph{Manual de
  computa{\c{c}}{\~a}o evolutiva e metaheur{\'\i}stica}, ser. Ensino.\hskip 1em
  plus 0.5em minus 0.4em\relax Imprensa da Universidade de Coimbra / Coimbra
  University Press, 2012. [Online]. Available:
  \url{https://books.google.com.br/books?id=9Di5CwAAQBAJ}
\BIBentrySTDinterwordspacing

\bibitem{diabetes}
\BIBentryALTinterwordspacing
J.~W. Smith, J.~Everhart, W.~Dickson, W.~Knowler, and R.~Johannes, ``Using the
  adap learning algorithm to forecast the onset of diabetes mellitus,''
  \emph{Proceedings. Symposium on Computer Applications in Medical Care}, p.
  261—265, November 1988. [Online]. Available:
  \url{https://europepmc.org/articles/PMC2245318}
\BIBentrySTDinterwordspacing

\bibitem{steel_2010}
M.~{Buscema}, S.~{Terzi}, and W.~{Tastle}, ``A new meta-classifier,'' in
  \emph{2010 Annual Meeting of the North American Fuzzy Information Processing
  Society}, 2010, pp. 1--7.

\bibitem{monks2}
S.~B. Thrun, J.~Bala, E.~Bloedorn, I.~Bratko, B.~Cestnik, J.~Cheng, K.~D. Jong,
  S.~Dzeroski, S.~E. Fahlman, D.~Fisher, R.~Hamann, K.~Kaufman, S.~Keller,
  I.~Kononenko, J.~Kreuziger, R.~Michalski, T.~Mitchell, P.~Pachowicz,
  Y.~Reich, H.~Vafaie, W.~V.~D. Welde, W.~Wenzel, J.~Wnek, and J.~Zhang, ``The
  monk's problems a performance comparison of different learning algorithms,''
  Tech. Rep., 1991.

\bibitem{phoneme}
J.-L. Voz, M.~Verleysen, P.~Thissen, and J.-D. Legat, ``A practical view of
  suboptimal bayesian classification with radial gaussian kernels,'' in
  \emph{Proceedings of the International Workshop on Artificial Neural
  Networks: From Natural to Artificial Neural Computation}, ser. IWANN
  ’96.\hskip 1em plus 0.5em minus 0.4em\relax Berlin, Heidelberg:
  Springer-Verlag, 1995, p. 404–411.

\bibitem{blood}
\BIBentryALTinterwordspacing
I.-C. Yeh, K.-J. Yang, and T.-M. Ting, ``Knowledge discovery on rfm model using
  bernoulli sequence,'' \emph{Expert Systems with Applications}, vol.~36, no.
  3, Part 2, pp. 5866 -- 5871, 2009. [Online]. Available:
  \url{http://www.sciencedirect.com/science/article/pii/S0957417408004508}
\BIBentrySTDinterwordspacing

\bibitem{ozone}
\BIBentryALTinterwordspacing
K.~Zhang and W.~Fan, ``Forecasting skewed biased stochastic ozone days:
  Analyses, solutions and beyond,'' \emph{Knowl. Inf. Syst.}, vol.~14, no.~3,
  p. 299–326, Mar. 2008. [Online]. Available:
  \url{https://doi.org/10.1007/s10115-007-0095-1}
\BIBentrySTDinterwordspacing

\bibitem{uci:2019}
\BIBentryALTinterwordspacing
D.~Dua and C.~Graff, ``{UCI} machine learning repository,'' 2017. [Online].
  Available: \url{http://archive.ics.uci.edu/ml}
\BIBentrySTDinterwordspacing

\bibitem{ilpd}
B.~V. Ramana, M.~S.~P. Babu, and N.~B. Venkateswarlu, ``A critical comparative
  study of liver patients from usa and india: An exploratory analysis,'' 2012.

\bibitem{sklearn_api}
L.~Buitinck, G.~Louppe, M.~Blondel, F.~Pedregosa, A.~Mueller, O.~Grisel,
  V.~Niculae, P.~Prettenhofer, A.~Gramfort, J.~Grobler, R.~Layton,
  J.~VanderPlas, A.~Joly, B.~Holt, and G.~Varoquaux, ``{API} design for machine
  learning software: experiences from the scikit-learn project,'' in \emph{ECML
  PKDD Workshop: Languages for Data Mining and Machine Learning}, 2013, pp.
  108--122.

\bibitem{red_wine}
\BIBentryALTinterwordspacing
P.~Cortez, A.~Cerdeira, F.~Almeida, T.~Matos, and J.~Reis, ``Modeling wine
  preferences by data mining from physicochemical properties.'' \emph{Decis.
  Support Syst.}, vol.~47, no.~4, pp. 547--553, 2009. [Online]. Available:
  \url{http://dblp.uni-trier.de/db/journals/dss/dss47.html#CortezCAMR09}
\BIBentrySTDinterwordspacing

\bibitem{gpu_kernel}
R.~Ballester-Ripoll, E.~G. Paredes, and R.~Pajarola, ``Sobol tensor trains for
  global sensitivity analysis,'' \emph{Reliability Engineering \& System
  Safety}, vol. 183, pp. 311--322, 2019.

\bibitem{breiman2001random}
L.~Breiman, ``Random forests,'' \emph{Machine learning}, vol.~45, no.~1, pp.
  5--32, 2001.

\bibitem{Goodfellow2015DeepL}
I.~G. Goodfellow, Y.~Bengio, and A.~C. Courville, ``Deep learning,''
  \emph{Nature}, vol. 521, pp. 436--444, 2015.

\bibitem{vapnik}
B.~E. Boser, I.~M. Guyon, and V.~N. Vapnik, ``A training algorithm for optimal
  margin classifiers,'' in \emph{Proceedings of the 5th Annual ACM Workshop on
  Computational Learning Theory}, pp. 144--152.

\bibitem{drucker1997support}
H.~Drucker, C.~J. Burges, L.~Kaufman, A.~J. Smola, and V.~Vapnik, ``Support
  vector regression machines,'' in \emph{Advances in neural information
  processing systems}, 1997, pp. 155--161.

\bibitem{Tikhonov:1963}
A.~N. Tikhonov, ``Solution of incorrectly formulated problems and the
  regularization method,'' \emph{Soviet Math. Dokl.}, vol.~4, pp. 1035--1038,
  1963.

\end{thebibliography}

\end{document}